\documentclass[conference]{IEEEtran}
\IEEEoverridecommandlockouts
\usepackage{cite}
\usepackage{amsmath,amssymb}
\usepackage{booktabs}
\usepackage{array}
\usepackage{graphicx}
\usepackage{url}
\usepackage{microtype}
\usepackage{tikz}
\usetikzlibrary{arrows.meta,positioning,fit,backgrounds}
\usepackage[T1]{fontenc}
\usepackage[utf8]{inputenc}

\title{Memory Canonicalization: A Framework and Benchmark for Cross-Model Drift in Persistent LLM Memory}

\author{\IEEEauthorblockN{Amit Vadnere}
\IEEEauthorblockA{Product Solutions Engineer\\
avadnere@google.com}
\and
\IEEEauthorblockN{Aishwarya Lonarkar}
\IEEEauthorblockA{Full Stack Developer\\
aishwaryalonarkar@gmail.com}}

\begin{document}
\maketitle

\begin{abstract}
Persistent memory for Large Language Models (LLMs) has matured rapidly: systems such as MemGPT/Letta, Mem0, and Zep now give agents tiered, temporally-aware, model-agnostic external storage, and the Model Context Protocol (MCP) now standardizes how any client reaches any memory server. What remains less addressed is a narrower but consequential problem: an identical stored memory object, retrieved by two different LLMs under otherwise identical conditions, is not guaranteed to be interpreted the same way---factually or emotionally. This paper proposes memory canonicalization: a write-time pipeline that detects ambiguity, conditional structure, and emotional loading in a raw memory object, and rewrites it into an explicit, structurally disambiguated canonical form---with emotional valence represented as a separate, explicit field rather than left for each consuming model to infer from tone---before it is committed to storage. We formalize the canonicalization pipeline, define a companion Cross-Model Semantic Drift / Emotional Consistency Score benchmark (CMSC-E), and report results from a three-arm pilot (raw vs. disambiguated-only vs. fully canonicalized memory, $n=176$ synthetic memory objects, three downstream model families, paired significance testing with correction for multiple comparisons). We find a significant uncorrected improvement in cross-model emotional consistency for fully canonicalized memory relative to raw memory (+0.050, 95\% bootstrap CI [0.013, 0.086], paired t-test $p=0.010$), but this result does not survive Bonferroni, Holm, or Benjamini--Hochberg correction across the six comparisons tested, and none of the factual-drift (CMSD) comparisons reach significance at any correction level. We report these results as exploratory rather than confirmatory, discuss likely sources of the null/underpowered findings, and outline the specific methodological changes---larger sample size, independent judge models, human-validated rendering, and a pre-registered follow-up---required before canonicalization's benefit can be claimed with confidence.
\end{abstract}


\section{INTRODUCTION}

Three years of rapid engineering progress have largely solved the
storage half of persistent LLM memory. MemGPT reframed an
LLM's context as virtual memory and gave agents
self-directed control over what stays in-context versus what gets paged
to external storage [4]; its production successor, Letta,
generalized this into a three-tier, model-agnostic runtime [5]. Mem0
removed the runtime-lock-in cost by offering memory as a bolt-on,
framework-agnostic service with a hybrid vector/graph/key-value backend
[6]. Zep, through its Graphiti engine, added the tier that both of
the above lacked: bi-temporal awareness, so a memory system knows not
just what is true but when it became true and when it was superseded
[7]. In parallel, the Model Context Protocol (MCP) -- now under
Linux Foundation governance and supported across Anthropic, OpenAI, and
Google DeepMind's platforms -- solved the transport
question: any client can now reach any compliant memory server through
one standard interface [13].

None of this solves a quieter problem sitting underneath all of it. A
memory object is not consumed in a vacuum -- it is retrieved and handed
to a specific model, which must interpret it, weigh it, and act on it.
Two different models, trained on different data with different alignment
procedures, are not guaranteed to interpret an identical object the same
way. This concern has begun to surface, independently, across several
unrelated corners of the 2026 literature: MemCollab found that naively
transferring memory between agents built on different backbone models
degrades task performance [17]; a cross-model continuity experiment
spanning ten model families found measurable tone and persona drift when
a shared external memory was handed off between models mid-conversation
[18]; and a dedicated emotional-consistency study found that three
different models each show degraded self-consistency under emotionally
loaded framing [24]. In every one of these papers, the contribution
stops at measurement -- or, for MemCollab, at a training-time fix
specific to reasoning-trajectory distillation that does not generalize
to stored personal-memory objects.

This paper proposes a mechanism that operates on the memory object
itself, before storage, to make it less prone to drift regardless of
which model later retrieves it: memory canonicalization. Section IV
formalizes this pipeline. Section V extends our CMSD metric with an
Emotional Consistency Score (ECS) into a combined CMSC-E benchmark.
Section VI walks through a hypothetical worked example to illustrate the
method's mechanics. Section VII reports actual results
from running the Section V-C three-arm pilot protocol end to end --
dataset generation, canonicalization, downstream querying across three
model families, and paired statistical testing with multiple-comparison
correction. We report these results plainly, including the ways they
complicate a simple confirmation of the paper's
hypothesis, and revise our limitations and future-work sections
accordingly.

\subsection*{Contributions}

\begin{itemize}
\item
  We introduce memory canonicalization, a write-time transformation for
  persistent LLM memory that resolves conditional and temporal ambiguity
  before storage.
\item
  We propose explicit emotional-field separation --- valence, intensity,
  and topic as distinct, validated fields --- as a first-class element
  of a persistent memory object's schema.
\item
  We introduce CMSC-E, to our knowledge the first benchmark to jointly
  measure semantic and emotional cross-model drift in retrieved memory
  objects under a controlled, three-arm rendering design.
\item
  We present the first empirical evaluation of this hypothesis,
  reporting paired, multiple-comparison-corrected results from a
  176-object pilot across three model families, including a null
  factual-drift result and an uncorrected-only emotional-consistency
  effect.
\end{itemize}

\section{RELATED WORK: A CHAIN OF PARTIAL SOLUTIONS}

This section is organized as a chain: each system below is presented as
a response to a limitation in the one before it, as a logical rather
than strictly historical progression.

\subsection{Attention and the Root Constraint}

Vaswani et al.'s transformer architecture [1] solved
flexible, content-based reasoning over a context window. Its unaddressed
limitation: nothing in the architecture persists once the window is
exceeded or a session ends.

\subsection{Retrieval-Augmented Generation}

Lewis et al. [2] solved the problem of grounding a model in
information beyond its context window via non-parametric retrieval.
Limitation: retrieval is stateless per call -- RAG has no concept of a
memory object persisting, updating, or being superseded across turns.

\subsection{Vector Databases}

FAISS [3] and its commercial successors made RAG viable at scale.
Limitation: the dense vector embedding carries no native concept of
confidence, provenance, or temporal validity; such structure has to be
bolted on as external metadata.

\subsection{MemGPT / Letta}

MemGPT [4], developed into the production runtime Letta [5],
solved persistence and self-directed memory management. Limitation:
nothing in the design checks whether two different models paging the
same archival memory object interpret it identically.

\subsection{Mem0}

Mem0 [6] broadened adoption by packaging memory as a bolt-on SDK with LLM-based fact extraction and a hybrid storage backend, making persistent memory easier to integrate across agent frameworks. Its own reported evaluation, however, identifies complex temporal reasoning as unresolved, and it does not test whether heterogeneous consuming models agree on the meaning of the extracted memory.

\subsection{Zep / Graphiti}

Zep [7], through Graphiti, advances the representation layer further with a bi-temporal knowledge graph that records when facts become valid and when they are superseded rather than silently overwriting them. This directly addresses memory evolution over time, but the system remains focused on storage and temporal semantics rather than consistency of interpretation across heterogeneous downstream models.

\subsection{Summarization, Reflection, and Knowledge-Graph
Extraction as Write-Time Transformations}

Mem0's fact extraction and Zep's graph
construction are themselves write-time transformations, which raises a
fair question: how does canonicalization differ from knowledge-graph
(KG) triplet extraction or summarization-based consolidation? Two
answers. First, Zhang et al. [15] show that continuously rewriting
an agent's memory through repeated LLM-driven
consolidation is actively degrading over many update cycles;
canonicalization is deliberately a single write-time pass with the raw
form always retained, which we assert by design as structurally
different, though this has not been tested under repeated
re-canonicalization. Second, KG triplet extraction structures entity
relations, a different axis than canonicalization targets: an accurately
extracted triplet can still leave conditional scope implicit, and does
not extract emotional valence as a first-class typed field. Multi-agent
KG-enrichment frameworks such as KARMA [14] are architecturally
close to our Phase 1/Phase 2 split, which we discuss directly in Section
VIII.

\subsection{Native Platform Memory}

ChatGPT Memory, Gemini's Personal Intelligence, and
Claude's Chat Memory solved deployment as a zero-setup
default. Limitation: each is siloed to its own vendor with no shared
schema.

\subsection{Model Context Protocol (MCP)}

MCP [13] standardizes transport and discovery between clients and memory-capable services; it does not specify the semantic representation of the memory objects being exchanged.

\subsection{MemCollab: The First Direct Measurement of Cross-Model
Memory Drift}

MemCollab [17] is the first system to name the underlying problem
directly: naively sharing memory across agents with different backbone
models degrades performance. Its fix, contrastive trajectory
distillation, requires paired reasoning trajectories from multiple
models solving the same task -- a requirement that does not exist for a
single stored personal fact with no paired trajectory to contrast
against, and it does not address emotional content.

\subsection{Cross-Model Persona and Temporal Continuity}

Yang et al.'s ARPM framework [18] measured tone
migration and persona-consistency degradation when switching backbone
models mid-conversation. Limitation: the object of study is persona/tone
continuity, not the content-level correctness or emotional
interpretation of a specific stored fact.

\subsection{Cross-Model Emotional Consistency}

Oram et al. [24] tested each of three models'{}
self-consistency with its own prior response under emotionally loaded
false-presupposition prompts, finding degraded consistency that worsens
with emotional intensity -- an observational, within-model finding that
motivates but does not establish the cross-model claim tested in this
paper.

\subsection{Shared-Memory Multi-Agent Infrastructure}

Systems such as KoMA [20], INMS [32], and Collaborative Memory
[21] solved safe multi-agent read/write access to a shared memory
pool. Limitation: none tests semantic equivalence across heterogeneous
model backends accessing the same object.

\subsection{Diagnostic Benchmarks for Memory-System Failure}

The Response Consistency Index [23] formalized a metric for
shared-versus-separate context configurations. Limitation: it is a
single-model or aggregate-system diagnostic that does not isolate
variance attributable to which model is consuming the memory.

\subsection{Surveys and Benchmarks}

Zhang et al. [8] and Wu et al. [9] survey agent memory
mechanisms broadly; LongMemEval [10] and LoCoMo [11] provide
standard recall-accuracy benchmarks. None measure cross-model agreement.

\subsection{Where the Chain Currently Ends}

Storage, deployment, and transport are solved; cross-model drift is
documented in task memory, persona/tone, and emotion from three
independent angles. What has not appeared is a write-time mechanism,
operating on individual stored memory objects, that is designed to improve the consistency of drift for
both factual and emotional content, together with a benchmark measuring
both dimensions under a controlled, statistically rigorous design. That
is the link this paper adds, and Section VII reports what happened when
we built and ran it.

\subsection{Comparative Summary}

Table I positions the present work against the systems discussed above
along five dimensions: whether the system persists memory at all,
whether it is temporally aware (i.e., tracks when a fact became true or
was superseded), whether it treats emotional content as a distinct,
typed representation, whether it has been evaluated for cross-model
consistency, and whether it applies a write-time canonicalization step.
No prior system reviewed here combines emotion-as-a-field with a
cross-model evaluation; this combination is the specific gap the present
work targets.

\begin{table*}[t]
\caption{Comparison of Persistent Memory and Cross-Model Drift Systems}
\label{tab:comparison}
\centering
\footnotesize
\setlength{\tabcolsep}{5pt}
\begin{tabular}{lccccc}
\toprule
\textbf{System} & \textbf{Persistence} & \textbf{Temporal} & \textbf{Emotion} & \textbf{Cross-Model} & \textbf{Canonicalization} \\\\
\midrule
MemGPT / Letta [4], [5] & Yes & No & No & No & No \\\\
Mem0 [6] & Yes & Partial & No & No & No \\\\
Zep / Graphiti [7] & Yes & Yes & No & No & No \\\\
MCP [13] & N/A & N/A & No & No & No \\\\
Native platform memory & Yes & No & No & No & No \\\\
MemCollab [17] & No & No & No & Partial & No \\\\
ARPM [18] & No & No & Partial & Partial & No \\\\
This work & Yes & Yes & Yes & Yes & Yes \\\\
\bottomrule
\end{tabular}
\end{table*}

\section{PROBLEM STATEMENT}

We decompose the problem into three sub-claims. Sub-claims 2 and 3 are
directly motivated by gaps identified in Sections II-L and II-O
respectively. Sub-claim 1 is a premise this paper introduces rather than
an established finding.

\begin{itemize}
\item
  Ambiguity in a raw memory object -- unresolved conditional
  structure and unresolved relative temporal references -- is a
  measurable driver of cross-model factual drift, distinct from
  task-reasoning drift and whole-conversation persona drift.
\item
  Emotional content embedded implicitly in a memory
  object's phrasing is a measurable driver of
  cross-model emotional-interpretation drift, extending [24] into
  the memory-persistence setting.
\item
  A write-time canonicalization step -- restructuring ambiguous
  conditions explicitly and representing emotional content as an
  explicit structured field -- is designed to improve the consistency of both forms of drift, without
  requiring paired multi-model trajectories or a whole-conversation
  governance layer.
\end{itemize}

\textbf{Scope.} We are not claiming that canonicalization resolves all sources of
cross-model disagreement. Reasoning-style drift, decoding-strategy
effects, and whole-conversation persona continuity are explicitly out of
scope and are treated as complementary problems addressed by other
systems in Section II's chain (in particular [17]
and [18]). The claim under test is narrower: that a specific,
identifiable class of drift -- drift attributable to unresolved
ambiguity and implicit emotional framing in the stored object itself --
is reducible by a write-time transformation of that object, independent
of which model later retrieves it.

\section{PROPOSED METHOD: MEMORY CANONICALIZATION}

\subsection{Design Principles}

Canonicalization is split into two phases with different statefulness
properties.

\begin{itemize}
\item
  Phase 1 (stateless, per-object) -- structural and emotional
  canonicalization. Runs once, when a memory object is created or
  updated, requiring no access to the rest of the memory store.
\item
  Phase 2 (stateful, store-aware) -- conflict and supersession
  reconciliation. Runs after Phase 1 and requires comparison against
  existing memory objects of the same or related type.
\item
  Make implicit structure explicit, rather than compress it away.
  Canonicalization restructures a memory object -- and where necessary
  lengthens it -- rather than summarizing it.
\item
  Separate emotional content from factual content into distinct
  fields, and into distinct objects where the underlying facts are
  themselves unrelated.
\item
  Preserve the original. The raw, pre-canonicalization object is
  retained for audit and provenance.
\end{itemize}

\subsection{Theoretical Motivation}

The design principles above rest on a specific conceptual claim about
why ambiguity in a stored memory object should produce cross-model
disagreement in the first place, and why explicit rewriting should
improve the consistency of it. We state that claim here without formal proof; the empirical
test of it is Section VII.

An ambiguous natural-language statement is not a single proposition but
a compact encoding of several candidate propositions, with the intended
one left to be recovered by whoever reads it. ``The deadline was moved,
but only if the client signs off'' admits at least two readings
depending on whether the client has, in fact, since signed off -- a
reading that is resolvable at write time but not from the
sentence's surface form alone. A model retrieving this
object later must pick one reading, implicitly, as part of generating a
response; which reading it picks depends on that model's
own priors, decoding behavior, and alignment-induced default
assumptions, none of which are shared across model families by
construction. The same holds for emotional tone: a sentence that merely
implies distress (``it is what it is, I guess'') requires the reading
model to infer both the presence and the intensity of an emotional state
from surface cues, a task on which different models -- with different
training data and different degrees of sensitivity to indirect affect --
are not guaranteed to converge. Prior work has shown that affective information need not remain implicit in natural-language interpretation and can instead be represented explicitly within computational models. Adam et al. [34], for example, formalize emotions in relation to beliefs, desires, expectations, and other mental attitudes using a modal logic grounded in the OCC theory of emotions. While their objective is formal reasoning about emotion rather than cross-model memory consistency, their work provides a precedent for explicitly representing affective content rather than leaving it entirely to downstream inference.

We describe this informally as the interpretation entropy of a memory
object: the number of materially different readings a competent reader
could assign to it, weighted by how plausible each reading is. An
unambiguous, fully scoped statement with an explicit emotional-valence
field has, by construction, an interpretation entropy at or near zero --
there is no remaining latitude for two models to resolve the object
differently, because the resolution has already been performed once, in
writing, at the point of storage. Canonicalization is, on this view, a
mechanism for shifting the point at which ambiguity is resolved from
many uncoordinated retrieval-time events (one per consuming model) to a
single write-time event whose output is then shared identically by every
subsequent reader. This does not guarantee that every model will act on
the resolved content identically -- downstream reasoning, decoding, and
alignment differences remain -- but it removes ambiguity in the stored
object itself as a distinct, independently addressable source of
disagreement, which is the narrower claim this paper tests.

\subsection{On Write-Time Interpretive Bias}

A single canonicalizing model resolving an ambiguous raw memory
necessarily makes a subjective interpretive choice, persisted as if it
were the object's canonical meaning -- replacing
distributed, retrieval-time disagreement between consuming models with a
single, centralized, write-time judgment call. Three partial frameworks
are available: (a) round-trip validation (Section IV-E, Steps 5/5b)
bounds how far the rewrite can drift from the raw
content's factual and emotional claims, though this
depends on the reliability of the judge models used, a concern that
Section VII's results make concrete; (b) a multi-model
ensemble canonicalizer is listed as future work; (c) the pipeline is
auditable by design, since raw content is always retained.

It follows that canonicalization is not intended to establish an
objective, model-independent ``true'' reading of an ambiguous memory --
no such ground truth is available for many of the conditional or
emotionally loaded statements this pipeline targets. What
canonicalization establishes instead is a single, deterministic
representation produced under a fixed, auditable policy, applied
identically regardless of which model later retrieves the object. The
relevant property is not correctness of interpretation but consistency
of interpretation across readers; a canonicalizer that is wrong but
consistent still removes the specific source of drift this paper
targets, though it may introduce a different, centralized bias in its
place -- a trade-off we consider explicitly rather than treat as
resolved (Section IX; Section X).

\subsection{Extended Memory Object Schema}

\{

\begin{quote}
"memory\_id": "string (uuid)",

"type": "enum: fact \textbar{} event \textbar{} preference \textbar{}
task",

"raw\_content": "string: original, unprocessed natural-language memory",

"canonical\_content": "array\textless string\textgreater: one or more
atomic, disambiguated statements derived from raw\_content",

"entities": "array\textless string\textgreater: named entities parsed in
Step 1",

"timestamp": "string (ISO-8601): capture time",

"valid\_from": "string (ISO-8601)",

"valid\_until": "string (ISO-8601) or null",

"confidence": "float, 0.0-1.0",

"sensitivity": "enum: low \textbar{} medium \textbar{} high",

"ambiguity\_level": "enum: none \textbar{} conditional \textbar{}
superseded \textbar{} conflicting",

"emotional\_valence": "enum: positive \textbar{} negative \textbar{}
neutral \textbar{} mixed \textbar{} null",

"emotional\_intensity": "float, 0.0-1.0, null iff emotional\_valence is
null",

"emotional\_topic\_flag": "string or null",

"canonicalization\_confidence": "float, 0.0-1.0, set by validators"
\end{quote}

\}

\subsection{Canonicalization Pipeline}

\begin{figure*}[t]
\centering
\begin{tikzpicture}[
  node distance=7mm and 7mm,
  box/.style={draw, rounded corners=2pt, align=center, minimum height=10mm, minimum width=25mm, fill=gray!8, font=\footnotesize},
  arrow/.style={-{Latex[length=2mm]}, thick},
  phase/.style={draw, dashed, rounded corners=3pt, inner sep=5pt}
]
\node[box] (raw) {Raw Memory};
\node[box, right=of raw] (parser) {Parser\\{\scriptsize NER + temporal normalization}};
\node[box, right=of parser] (detect) {Ambiguity / Emotion\\Detection};
\node[box, right=of detect] (canon) {Canonicalizer\\{\scriptsize rewriting + object splitting}};
\node[box, below=of canon] (emotion) {Emotion Extraction};
\node[box, left=of emotion] (validate) {Validators\\{\scriptsize factual + emotional}};
\node[box, left=of validate] (conflict) {Conflict Resolution\\{\scriptsize Phase 2, store-aware}};
\node[box, left=of conflict] (store) {Memory Store};
\draw[arrow] (raw) -- node[above, font=\scriptsize]{Step 1} (parser);
\draw[arrow] (parser) -- node[above, font=\scriptsize]{Step 2} (detect);
\draw[arrow] (detect) -- node[above, font=\scriptsize]{Step 3} (canon);
\draw[arrow] (canon) -- node[right, font=\scriptsize]{Step 4} (emotion);
\draw[arrow] (emotion) -- node[above, font=\scriptsize]{Steps 5/5b} (validate);
\draw[arrow] (validate) -- node[above, font=\scriptsize]{Step 6} (conflict);
\draw[arrow] (conflict) -- node[above, font=\scriptsize]{Step 7} (store);
\begin{scope}[on background layer]
  \node[phase, fit=(raw)(parser)(detect)(canon)(emotion)(validate), label={[font=\scriptsize]above:Phase 1: stateless, per-object}] {};
  \node[phase, fit=(conflict), label={[font=\scriptsize]below:Phase 2: store-aware}] {};
\end{scope}
\end{tikzpicture}
\caption{Canonicalization pipeline architecture. Steps 1--5b (Phase 1) are stateless and run per object; Step 6 (Phase 2) is store-aware and runs after Phase 1.}
\label{fig:pipeline}
\end{figure*}
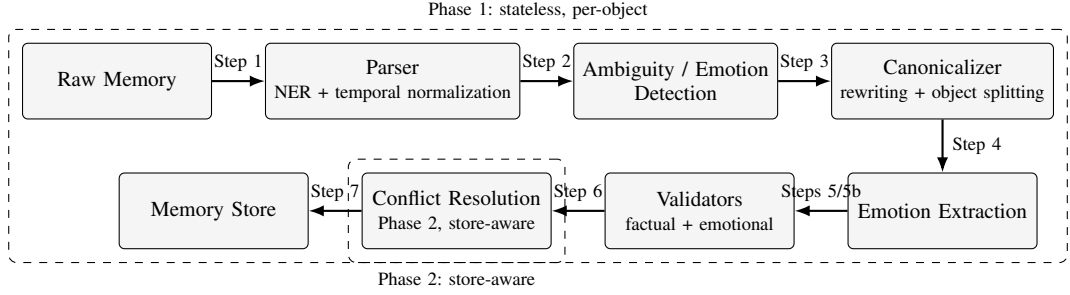

Steps 1--5 constitute Phase 1; Step 5b closes an emotional-validation
gap; Step 6 is Phase 2; Step 7 is storage. Fig.~\ref{fig:pipeline} summarizes the
resulting data flow.

Model roles. Four distinct model/module roles are used so that no single
model simultaneously parses, writes, judges, and validates the same
content: (i) a deterministic parser (NER plus a temporal normalizer)
performs Step 1; (ii) a canonicalizer LLM performs Steps 2--4; (iii) a
factual NLI validator performs Step 5; (iv) an emotional validator,
ideally a discriminative classifier architecturally distinct from the
canonicalizer, performs Step 5b.

\textbf{Step 1 -- Structural parse.} Extracts entities and
resolves explicit/implicit conditions and relative time expressions into
structured intermediate output.

\textbf{Step 2 -- Ambiguity and emotional-loading detection.}
Flags the object along two axes: unresolved conditional structure, and
emotionally loaded language whose sentiment is currently only implicit.

\textbf{Step 3 -- Canonical rewriting and object splitting.}
Produces one or more atomic statements making every condition, scope,
and temporal reference explicit; splits substantively unrelated merged
facts into separate objects.

\textbf{Step 4 -- Emotional-field extraction.} Produces
emotional\_valence, emotional\_intensity, and emotional\_topic\_flag for
the already-split object(s) from Step 3.

\textbf{Step 5 -- Factual round-trip validation.} A precision
check (canonical content must be entailed by raw content plus timestamp)
and a coverage check (every atomic claim in the raw content must be
entailed by the union of the resulting canonical objects), run by a
dedicated NLI validator.

\textbf{Step 5b -- Emotional-field plausibility validation.} An
independent emotional validator re-scores the raw content for valence
and intensity; disagreement beyond a threshold routes the object to
human review.

\textbf{Step 6 -- Conflict and supersession reconciliation (Phase
2).} Compares Phase-1 output against existing objects of the same type
and overlapping entities; sets ambiguity\_level to superseded or
conflicting as appropriate, non-destructively.

\textbf{Step 7 -- Storage.} Both raw\_content and the canonical
object(s) are stored; retrieval returns canonical content by default.

\subsection{Complexity Considerations}

Phase 1 is a bounded, per-object pipeline: each of Steps 1--5b runs once
per memory object and processes only that object's
content, so Phase 1 cost is linear in the length of the raw memory
object and independent of the size of the memory store. In practice this
cost is dominated by the canonicalizer and validator LLM calls (Steps
2--5b), so wall-clock latency and token overhead scale with the number
of LLM invocations per object (four in the pipeline as specified) rather
than with store size, and can be amortized because Phase 1 runs once at
write time rather than on every retrieval. Phase 2 (Step 6) is
store-aware by design: its cost depends on the candidate-retrieval
mechanism used to find existing objects of the same type and overlapping
entities (e.g., an index lookup versus a full-store scan), and is
therefore implementation-dependent rather than fixed by the method
itself. We do not report measured latency or dollar-cost figures for the
Section VII pilot, since the pilot's purpose was to test
drift reduction rather than production throughput; characterizing
end-to-end latency and cost under realistic store sizes is listed as
future work (Section XI).

\textbf{V. THE CMSC-E BENCHMARK: EXTENDING CMSD WITH EMOTIONAL
CONSISTENCY}

\subsection{CMSD (Recap)}

For a memory object and downstream prompt, response claim sets from each
model under test are extracted, and pairwise bidirectional entailment
agreement is computed and averaged, producing a score in {[}0, 1{]} per
memory object.

\subsection{Emotional Consistency Score (ECS)}

For a memory object carrying non-null emotional\_valence, each
downstream model's response is scored along three axes
using two different judges: valence and intensity are scored by the
emotional-validator classifier; topic-handling (does the model
reference, avoid, or misjudge the flagged topic) is scored by the
factual-validator NLI model, reused for entailment between the response
and emotional\_topic\_flag. The three per-axis scores are combined into
ECS via unweighted mean by default -- a simplifying assumption we flag
in Section X.

\subsection{Experimental Protocol -- Three-Arm Design}

An earlier two-arm design (raw prose vs. structured JSON) confounds
content (whether ambiguity is resolved and emotion made explicit) with
format (whether the model receives free text or structured data). We use
a three-arm design with a fixed rendering rule instead.

Rendering rule. Every memory object is rendered into the downstream
prompt as flat natural-language prose regardless of condition -- the
model under test never sees field names directly. Continuous fields
(emotional intensity) use a bank of human-authored, semantically
equivalent phrasings sampled per band, so surface repetition does not
confound the comparison.

\begin{itemize}
\item
  Arm 1 -- Raw. The unprocessed raw\_content, as-is.
\item
  Arm 2 -- Disambiguated-only (ablation). The Step 3 structural rewrite,
  without Step 4 emotional-field extraction.
\item
  Arm 3 -- Fully canonicalized. Steps 3 and 4 together.
\end{itemize}

Comparing Arm 1 to Arm 2 isolates the effect of structural
disambiguation alone; comparing Arm 2 to Arm 3 isolates the effect of
explicit emotional extraction specifically.

\section{ILLUSTRATIVE WORKED EXAMPLE}

(Hypothetical, for methodology demonstration only -- distinct from the
actual pilot data reported in Section VII.) Consider: ``User mentioned
their dog passed away last month, seemed pretty upset about it -- also
said the Tuesday appointment reschedule was fine, no issue there.'' This
sentence merges two unrelated pieces of content with very different
emotional weight. Under Step 3, the raw object is split into two
canonical memory objects; Step 4 computes emotional annotation
independently per object.

Object 1 (the loss) receives emotional\_valence: ``negative'',
emotional\_intensity: 0.7, emotional\_topic\_flag: ``pet loss / grief''.

Object 2 (the reschedule) receives emotional\_valence: ``neutral'',
emotional\_intensity: 0.05 (non-null, since ``neutral'' is an assessed
rating, distinct from null meaning ``not assessed''). With the emotional
register stated explicitly and the two facts stored separately, the
prediction under this paper's hypothesis was that models
would converge more consistently on how to reference the loss. Section
VII reports whether this prediction held.

\section{PILOT RESULTS}

This section reports results from running the Section V-C three-arm
protocol to completion. All results below use real API calls against
three production models; no numbers in this section are simulated or
illustrative.

\subsection{Dataset and Setup}

We generated 176 synthetic memory objects, stratified across 16 strata
(4 memory types $\times$ 2 ambiguity levels $\times$ 2 emotion states), using
template-based generation with randomized names, dates, and phrasing to
avoid a single fixed surface form per stratum. The sample size was chosen to provide balanced coverage across the experimental strata while keeping the resulting cross-model evaluation computationally tractable. We treated 176 objects as a pilot-scale sample sufficient for estimating within-dataset paired effects rather than as a claim of population-level coverage; evaluating substantially larger samples is left for future work. Each object was processed
through the full Phase 1 pipeline (Section IV-E, Steps 1--5b): a
deterministic parser (regex-based, with an optional spaCy/dateparser
backend verified to produce consistent entity and relative-date
extraction on a manual sample of pipeline output), a canonicalizer LLM,
a factual NLI validator, and an emotional validator. All 176 objects
completed Phase 1 with zero pipeline errors.

Each object's three arms (Section V-C) were rendered as
flat prose and paired with two human-authored downstream prompts drawn
from a fixed, type-specific template bank. Each (object, arm, prompt)
triple was sent in parallel to three models under test: one
closed-weight frontier model, one closed-weight smaller/faster model,
and one open-weight model, spanning three separate model families as the
protocol specifies. CMSD was computed via pairwise claim-entailment
agreement across the three models'{} responses; ECS was
computed for the subset of objects carrying non-null emotional content,
via the valence/intensity/topic-handling procedure in Section V-B.

Of the 176 generated objects, 172 yielded a valid CMSD score in every
arm; four were excluded because, after our exclusion policy for
timed-out or failed downstream-model calls (a call that fails is
excluded from scoring rather than treated as a real response), fewer
than two valid model responses remained for at least one arm/prompt
combination. Of the 172, 86 carried non-null emotional content per
Phase-1 canonicalization and were eligible for ECS scoring.

\subsection{Statistical Methodology}

Because the same 172 (respectively 86) objects were scored under all
three arms, arm comparisons are paired observations, not independent
samples. For each of the three arm-pairs (Arm 1 vs. 2, Arm 2 vs. 3, Arm
1 vs. 3) and each of the two metrics (CMSD, ECS), we report: the paired
mean difference; a paired t-test; a Wilcoxon signed-rank test
(non-parametric, robust to the non-normal, bounded-{[}0, 1{]}
distribution CMSD/ECS scores are likely to have); and a 10,000-resample
bootstrap 95\% confidence interval on the mean difference, which
resamples object pairs jointly to preserve pairing. This yields six
hypothesis tests in total (three arm-pairs $\times$ two metrics), which we
treat as a multiple-comparisons family and correct accordingly using
Bonferroni, Holm, and Benjamini--Hochberg procedures at family-wise $\alpha$ =
0.05.

\subsection{Results}

Table II reports the full set of six comparisons.

\emph{\textbf{CMSD (factual drift).}} None of the three CMSD comparisons
approach conventional significance (p\_t $\geq$ 0.46 in all three cases; all
bootstrap CIs comfortably include zero). The point estimate for
disambiguation (Arm 1 $\rightarrow$ 2) is positive and in the predicted direction
(+0.009), but small relative to its uncertainty. The point estimate for
emotional extraction (Arm 2 $\rightarrow$ 3) is negative ($-$0.004) -- the only one of
the six comparisons pointing against the paper's
predicted direction, though its magnitude is negligible and
statistically indistinguishable from zero. Sub-claim 1 -- that ambiguity
in a raw memory object is a measurable driver of cross-model factual
drift correctable by disambiguation -- is not supported by this pilot at
this sample size.

\emph{\textbf{ECS (emotional drift), uncorrected.}} The overall
comparison (Arm 1 $\rightarrow$ 3) shows a positive, nominally significant effect:
+0.050 (95\% CI {[}0.013, 0.086{]}, p\_t = 0.010, p\_w = 0.006). Both
sub-effects trend in the same positive direction (disambiguation +0.018,
p\_t = 0.294; emotional extraction +0.031, p\_t = 0.103) without
individually reaching significance -- the emotional-extraction
sub-effect is closer to conventional significance than the
disambiguation sub-effect, consistent with (though not proof of) the
paper's Sub-claim 2/3 framing that explicit emotional
annotation specifically, not disambiguation alone, drives the
emotional-consistency benefit.

\emph{\textbf{Correction for multiple comparisons.}} Across the six
tests conducted, sorted p-values are \{0.010, 0.103, 0.294, 0.463,
0.664, 0.721\}. The Bonferroni-corrected threshold for the
smallest-ranked test is $\alpha$/6 = 0.0083; the smallest observed p-value,
0.010, exceeds this threshold and is therefore not significant under
Bonferroni correction. Holm's step-down procedure fails
at the first step for the same reason (the same threshold applies to the
smallest p-value at rank 1), and therefore rejects nothing.
Benjamini--Hochberg, applied at each rank i against threshold (i/6) $\times$
0.05, also fails at rank 1 (0.010 \textgreater{} 0.0083), and since BH
is a step-up procedure that requires the largest p-value below its own
rank-specific threshold, no test in this family is flagged as a
discovery. Under all three standard correction procedures, none of the
six comparisons in this pilot reach statistical significance.

\begin{table*}[t]
\caption{Paired Arm Comparisons, CMSD ($n=172$) and ECS ($n=86$)}
\label{tab:results}
\centering
\footnotesize
\setlength{\tabcolsep}{4pt}
\begin{tabular}{llrrrrrr}
\toprule
\textbf{Metric} & \textbf{Comparison (A $\rightarrow$ B)} & \textbf{Mean A} & \textbf{Mean B} & \textbf{Diff} & \textbf{95\% CI} & \textbf{$p_t$} & \textbf{$p_w$} \\\\
\midrule
CMSD & Arm 1 $\rightarrow$ 2 (disambiguation) & 0.7301 & 0.7388 & +0.0088 & [$-0.014$, 0.033] & 0.463 & 0.729 \\\\
CMSD & Arm 2 $\rightarrow$ 3 (emotional extraction) & 0.7388 & 0.7349 & $-0.0039$ & [$-0.022$, 0.013] & 0.664 & 0.573 \\\\
CMSD & Arm 1 $\rightarrow$ 3 (overall) & 0.7301 & 0.7349 & +0.0048 & [$-0.022$, 0.032] & 0.721 & 0.931 \\\\
ECS & Arm 1 $\rightarrow$ 2 (disambiguation) & 0.6818 & 0.7002 & +0.0184 & [$-0.015$, 0.052] & 0.294 & 0.292 \\\\
ECS & Arm 2 $\rightarrow$ 3 (emotional extraction) & 0.7002 & 0.7316 & +0.0314 & [$-0.006$, 0.068] & 0.103 & 0.079 \\\\
ECS & Arm 1 $\rightarrow$ 3 (overall) & 0.6818 & 0.7316 & +0.0498 & [0.013, 0.086] & 0.010 & 0.006 \\\\
\bottomrule
\end{tabular}
\vspace{1mm}
\begin{minipage}{0.98\textwidth}
\footnotesize \textit{Note:} CI = 95\% bootstrap confidence interval on the mean difference (Arm B $-$ Arm A). $p_t$ = paired t-test; $p_w$ = Wilcoxon signed-rank. Bonferroni/Holm/BH-corrected threshold at family-wise $\alpha=0.05$ across all six tests is $p<0.0083$ for the smallest-ranked test; none of the six tests reach this threshold.
\end{minipage}
\end{table*}

\subsection{Effect Direction and a Note on Interim Analysis}

Five of the six point estimates are directionally positive (consistent
with the paper's hypothesis); only the CMSD Arm 2 $\rightarrow$ 3
comparison is negative, and its magnitude ($-$0.004) is well within noise.
This directional consistency is weak, non-dispositive evidence and
should not be over-read given that none of the individual effects clear
even an uncorrected significance threshold except the one overall ECS
comparison, which itself does not survive correction.

During pilot execution, we computed and inspected running means at
several intermediate sample sizes (n = 14, 19, 33, 43, 108) before
reaching the final n = 172/86. At smaller n, several of these interim
snapshots showed the disambiguation effect on CMSD as the visually
larger and more consistent-looking trend, with the emotional-extraction
effect appearing flat or even negative on both metrics at one checkpoint
(n = 43). By n = 108 the pattern had partially reversed, with ECS
beginning to favor Arm 3 over Arm 2. The final, fully powered, properly
paired-and-corrected analysis in Table II does not confirm the CMSD
effect that appeared most robust at smaller n, and only weakly (and not
multiple-comparison-robustly) supports the ECS effect. This is offered
as a direct, first-hand illustration of exactly the risk the
paper's own Section V-C already warns about: unweighted,
uncorrected inspection of intermediate means during data collection is
not a substitute for a pre-specified, fully powered, corrected analysis,
and can point in a different direction than the final result.

\subsection{An Approximate Power Analysis}

Treating the bootstrap standard errors implied by Table
II's confidence-interval widths (approx. 0.009--0.019
across the six comparisons) as representative, the minimum effect size
detectable at 80\% power and two-sided $\alpha$ = 0.05 with n = 172 is
approximately 2.8 $\times$ SE $\approx$ 0.033 for the CMSD comparisons -- roughly 3--4$\times$
larger than the disambiguation effect actually observed (0.009).
Achieving 80\% power to detect an effect of the observed magnitude at
this noise level would require on the order of 10--15$\times$ the current
sample size. We report this as an approximate, illustrative calculation
rather than a precise power computation, but it supports a specific
interpretation: the null CMSD result in this pilot is at least as
consistent with ``a true effect too small to detect at n = 172 given
this measurement approach's noise floor'' as it is with
``no true effect,'' and the two are not distinguishable without a
substantially larger sample.

\subsection{Threats to Validity}

Three implementation choices, made for pilot tractability, bear directly
on how the Section VII results should be read, and are distinct from the
design-level limitations already discussed in Sections IV-C and IX.

\emph{\textbf{Judge-model non-independence.}} The canonicalizer (Steps
2--4), the factual validator (Step 5), and the emotional-validator
stand-in (Step 5b) were all instruction-following LLMs from the same
model family, rather than the architecturally distinct discriminative
classifier the method specifies for Step 5b (Section IV-E). Because the
same family also supplied one of the three downstream models under test,
there is a plausible -- though unmeasured -- risk of correlated
judgments: a same-family judge scoring a same-family
model's response, or a same-family canonicalizer
producing text a same-family downstream model is disproportionately
well-calibrated to interpret. This would tend to inflate, not deflate,
apparent cross-model agreement for canonicalized arms, which is the
opposite direction needed to explain the null CMSD result, but it
complicates interpretation of the positive (if uncorrected) ECS result.

\emph{\textbf{Unvalidated rendering bank.}} Section V-C specifies that
the phrasing bank used to render continuous emotional-intensity values
into prose should be validated by human raters for naturalness and
correct intensity-band boundaries before the pilot is run. This
validation step was not performed for the results in this section. If
certain phrasings in the bank are systematically more or less ambiguous
than others, this could introduce measurement noise that specifically
affects the ECS metric, since ECS is the only score sensitive to this
rendering.

\emph{\textbf{No pre-registration.}} The three-arm comparisons, the
choice of paired tests, and the multiple-comparison correction were
specified and applied after data collection, not pre-registered
beforehand. We believe the analysis plan reported here is the natural,
specification-following implementation of Section V-C's
protocol rather than a post hoc search for significance, and we report
all six tests rather than a selected subset, but a pre-registered
replication remains the appropriate next step before treating any result
here as confirmatory.

\subsection{Interpreting the Null and Weak Results}

The CMSD null result admits more than one reading, and we consider it
important not to collapse these into a single narrative. One reading,
developed in Section VII-E, is that the pilot is simply underpowered for
an effect of the size actually observed. A second, equally plausible
reading is that ambiguity resolution in the stored memory object is not,
in fact, the dominant source of cross-model factual disagreement that
Sub-claim 1 assumes it to be. Architectural differences between model
families -- differences in pretraining data, alignment procedure, and
decoding strategy -- may simply contribute more variance to how a model
answers a downstream question than the phrasing of the retrieved memory
object does; if so, no amount of write-time disambiguation of the memory
object alone would be expected to close that gap, because the
disagreement would originate downstream of retrieval rather than in the
object itself. Our design cannot distinguish these two readings, because
we did not vary model-family composition or measure a
within-model-family baseline level of disagreement independent of memory
phrasing. We regard this as the most important open question the null
CMSD result raises, and we return to it directly in Section XI.

The ECS result is weaker evidence for the paper's
hypothesis than a single uncorrected p-value might suggest, and we do
not treat it as establishing Sub-claim 2 or 3. It is, at most,
consistent with those sub-claims: the direction of every ECS sub-effect
matches the prediction, the overall effect is nominally significant
before correction, and the effect is numerically larger for emotional
extraction than for disambiguation alone -- the pattern one would expect
if explicit emotional-field extraction specifically, rather than
structural rewriting in general, were doing the work. But the same
result is also consistent with a smaller true effect inflated by the
judge-independence and rendering-validation gaps documented above. We
report it as suggestive and flag the specific follow-up (Section XI)
required to adjudicate between these readings.

\section{RELATIONSHIP TO EXISTING SYSTEMS}

Canonicalization is designed as a pre-processing layer compatible with,
not competitive against, every system in Section II's
chain. It is explicitly narrower in scope than MemCollab [17] and
the persona-governance framework [18]: it does not attempt to fix
reasoning-trajectory transfer or whole-conversation tone continuity, and
should be read as complementary to both. On the question of whether
stateless-extraction-plus-stateful-reconciliation is itself a novel
architectural pattern: it is not, and we do not claim it is.
Mem0's hybrid backend and KARMA's
extraction-plus-conflict-resolution pipeline [14] already separate
these passes in essentially this shape. What we claim as contribution is
narrower: (a) the specific claim that emotional content needs its own
explicit, separately-validated field, which no system reviewed in
Section II implements; and (b) the CMSC-E benchmark and three-arm,
statistically corrected protocol (Section V, Section VII) for testing
whether that design choice measurably is designed to improve the consistency of cross-model drift -- an
empirical instrument, now exercised end to end, that the architectural
pattern alone does not provide.

\section{ETHICAL CONSIDERATIONS}

Storing an algorithmic inference about a user's
emotional state as a persistent, structured, machine-readable field is a
materially different privacy commitment than storing a factual
preference. Three issues are worth separating. Consent: a user
disclosing a fact in conversation has not necessarily consented to that disclosure being converted into a permanent, queryable emotional-intensity score; explicit consent for emotional-field extraction, separate from
consent to factual memory storage, is a reasonable default.
Reducibility: a discrete valence/intensity pair is a lossy reduction of
a nuanced human state, and risks being treated as more authoritative
than it is, particularly if a downstream reviewer treats a high
emotional\_intensity score as clinical or diagnostic signal. Right to
correction and deletion: a user should have a clear path to review,
correct, or delete inferred emotional annotations independent of the
factual content they attach to. The pilot reported in Section VII used
entirely synthetic, template-generated data with no real user
disclosures, which sidesteps these consent concerns for this specific
run but does not substitute for implementing the safeguards above before
canonicalization is applied to real user memory.

\section{LIMITATIONS}

\begin{itemize}
\item
  The core hypothesis is not confirmed by this pilot. Sub-claim 1
  (disambiguation is designed to improve the consistency of factual drift) is not supported at n = 172.
  Sub-claim 3's emotional-consistency prediction shows a
  positive, uncorrected-significant effect that does not survive
  correction for the six comparisons tested. This
  paper's contribution at this stage is the benchmark,
  the pipeline, and a rigorously reported null-to-weak empirical result
  -- not a confirmed positive finding.
\item
  The pilot is underpowered for the effect sizes observed. Section
  VII-E's approximate power calculation suggests the
  current sample size would need to grow by roughly an order of
  magnitude to reliably distinguish the observed point estimates from
  zero, if those point estimates reflect a true (if small) effect.
\item
  Judge-model independence is not achieved. Step 5b uses an LLM-based
  stand-in rather than the architecturally distinct discriminative
  classifier the method specifies, and the canonicalizer, factual
  validator, and one downstream model under test share a model family, a
  threat to validity discussed in Section VII-F.
\item
  The rendering bank was not human-validated before use, contrary to the
  protocol specified in Section V-C.
\item
  The supplementary superseded/conflicting set (30 paired objects,
  Section V-C) was constructed but not scored in this version of the
  paper; Phase-2 reconciliation's effect on drift
  remains untested.
\item
  The dataset is synthetic, template-generated with a bounded vocabulary
  of names, cities, and scenario types. Effect sizes and even effect
  directions measured here may not transfer to real, unstructured
  conversational memory.
\item
  No pre-registration. The statistical analysis plan, while
  specification-following, was finalized after data collection.
\item
  This does not address memory lifecycle security (poisoning, consent,
  cross-session contamination), flagged as a distinct open problem in
  [16], nor whether reasoning-style drift is removable by rewriting
  content alone, as raised by MemCollab and [18].
\item
  The differentiation from summarization-based consolidation (Section
  II-G) remains argued by design, not demonstrated empirically under
  repeated re-canonicalization.
\item
  Ethical safeguards (Section IX) are not implemented in the pipeline
  and remain a required extension before deployment on real user data.
\end{itemize}

\section{FUTURE WORK}

\begin{itemize}
\item
  Scale the pilot by an order of magnitude and pre-register the analysis
  plan (comparisons, tests, and correction method) before data
  collection, directly addressing Section VII-E's power
  finding.
\item
  Decouple judge models from the canonicalizer and from the models under
  test. Replace the Step 5b LLM stand-in with a dedicated discriminative
  classifier, and ensure no model family used as a judge or
  canonicalizer is also a model under test, to close the validity gap
  identified in Section VII-F.
\item
  Human-validate the rendering bank for naturalness and correct
  intensity-band boundaries, as originally specified in Section V-C,
  before the next data collection round.
\item
  Run the supplementary superseded/conflicting set (Section V-C) to test
  Phase-2 reconciliation's effect on drift, which
  remains completely untested in this version.
\item
  Replace synthetic data with real (consented, anonymized)
  conversational memory to test whether the null CMSD result and the
  weak ECS result are artifacts of the template-generated
  dataset's limited linguistic variety.
\item
  Investigate the negative point estimate on the emotional-extraction /
  CMSD comparison specifically -- even though not statistically
  distinguishable from zero here, a larger, well-powered study should
  determine whether explicit emotional-field rendering introduces any
  factual-agreement cost, since this would be a meaningful trade-off to
  document either way.
\item
  Implement and test the multi-model ensemble canonicalizer proposed in
  Section IV-C as a direct framework for the write-time interpretive
  bias and judge-independence concerns raised in Sections IV-C and
  VII-F.
\item
  Extend the emotional-field schema toward discrete-emotion taxonomies
  if a properly powered follow-up indicates the coarser
  valence/intensity representation is insufficient.
\item
  Evaluate canonicalization under heterogeneous canonicalizers -- i.e.,
  re-run the Section VII protocol with the canonicalizer role filled by
  different model families across replications -- to quantify how much
  of the write-time output, and any resulting drift reduction, is
  attributable to the specific canonicalizer model chosen rather than to
  canonicalization as a method.
\end{itemize}

\section{CONCLUSION}

This paper set out to test whether write-time memory canonicalization --
resolving ambiguity and making emotional content explicit before a
memory object is stored -- measurably is designed to improve the consistency of cross-model drift, factual
and emotional, when that memory is later retrieved by different LLMs. We
built the full pipeline the method specifies, generated a stratified
176-object pilot dataset, ran it end to end against three model families
under a three-arm design, and analyzed the results with paired
significance testing corrected for the six comparisons the protocol
entails. The honest result is a mixed and largely null one:
disambiguation alone does not produce a statistically detectable
reduction in cross-model factual drift at this sample size (n = 172),
and while fully canonicalized memory shows a nominally significant
improvement in cross-model emotional consistency relative to raw memory
(+0.050, p = 0.010 uncorrected), this single positive finding does not
survive correction for multiple comparisons and should be read as
suggestive rather than confirmed. We consider this an informative
negative-to-weak result rather than a failed paper: the pipeline runs
correctly end to end, the benchmark discriminates directionally
consistent (if mostly non-significant) effects, and the approximate
power analysis in Section VII-E gives a concrete, actionable target --
roughly an order of magnitude more data -- for the next iteration to
determine whether the observed point estimates reflect a real, small
effect or genuine noise. Until that follow-up is run, memory
canonicalization should be understood as a plausible, mechanistically
motivated proposal with a rigorously reported but not yet statistically
established benefit, and we report it this way deliberately, consistent
with this paper's own repeated caution against treating
exploratory pilot data as confirmatory. Regardless of whether future experiments ultimately validate or refute the hypothesis, the primary contribution of this work is CMSC-E together with a reproducible evaluation framework for studying cross-model drift in persistent LLM memory. Memory Canonicalization is presented as a testable framework that enables this evaluation and provides a foundation for future investigation.

\section*{Code Availability}

The official implementation of the Memory Canonicalization framework, the CMSC-E benchmark, and the experimental pipeline is publicly available at:

\url{https://github.com/aishwaryalonarkar/cross-model-memory-canonicalization}

\section*{REFERENCES}

[1] A. Vaswani, N. Shazeer, N. Parmar, J. Uszkoreit, L. Jones, A. N.
Gomez, Ł. Kaiser, and I. Polosukhin, ``Attention Is All You Need,''
NeurIPS, 2017.

[2] P. Lewis, E. Perez, A. Piktus, et al., ``Retrieval-Augmented
Generation for Knowledge-Intensive NLP Tasks,'' NeurIPS, 2020.

[3] J. Johnson, M. Douze, and H. Jégou, ``Billion-Scale Similarity
Search with GPUs,'' IEEE Transactions on Big Data, 2019.

[4] C. Packer, S. Wooders, K. Lin, V. Fang, S. G. Patil, and J. E.
Gonzalez, ``MemGPT: Towards LLMs as Operating Systems,''
arXiv:2310.08560, 2023.

[5] Letta AI, ``Letta: A Framework for Stateful LLM Agents (formerly
MemGPT),'' https://letta.ai, 2024--2026.

[6] P. Chhikara, D. Khant, S. Aryan, T. Singh, and D. Yadav, ``Mem0:
Building Production-Ready AI Agents with Scalable Long-Term Memory,''
arXiv:2504.19413, 2025.

[7] P. Rasmussen et al., ``Zep: A Temporal Knowledge Graph
Architecture for Agent Memory,'' arXiv:2501.13956, 2025.

[8] Z. Zhang, Q. Dai, L. Chen, Z. Jiang, R. Li, J. Zhu, X. Chen, Y.
Xie, Z. Dong, and J.-R. Wen, ``A Survey on the Memory Mechanism of Large
Language Model-based Agents,'' arXiv:2404.13501, 2024; ACM Transactions
on Information Systems, 2025.

[9] Y. Wu, S. Liang, C. Zhang, Y. Wang, Y. Zhang, H. Guo, R. Tang,
and Y. Liu, ``From Human Memory to AI Memory: A Survey on Memory
Mechanisms in the Era of LLMs,'' arXiv:2504.15965, 2025.

[10] D. Wu, H. Wang, W. Yu, Y. Zhang, K.-W. Chang, and D. Yu,
``LongMemEval: Benchmarking Chat Assistants on Long-Term Interactive
Memory,'' arXiv:2410.10813, 2024.

[11] A. Maharana, D.-H. Lee, S. Tulyakov, M. Bansal, F. Barbieri,
and Y. Fang, ``Evaluating Very Long-Term Conversational Memory of LLM
Agents,'' Proc. ACL, 2024.

[12] W. Xu, Z. Liang, K. Mei, H. Gao, J. Tan, and Y. Zhang, ``A-MEM:
Agentic Memory for LLM Agents,'' arXiv:2502.12110; NeurIPS, 2025.

[13] Anthropic, ``Model Context Protocol Specification,'' 2024
(Linux Foundation governance transition, Dec. 2025).

[14] Y. Lu and J. Wang, ``KARMA: Leveraging Multi-Agent LLMs for
Automated Knowledge Graph Enrichment,'' arXiv:2502.06472, 2025; NeurIPS,
2025.

[15] D. Zhang, Y. Lin, Z. Wu, Y. Sun, B. Li, D. Li, and H. Peng,
``Useful Memories Become Faulty When Continuously Updated by LLMs,''
arXiv:2605.12978, 2026.

[16] Z. Lin, X. Hao, R. Fu, S. Cui, K. Chen, C. Li, Z. Li, and F.
Xiong, ``A Survey on Long-Term Memory Security in LLM Agents: Attacks,
Defenses, and Governance Across the Memory Lifecycle,''
arXiv:2604.16548, 2026.

[17] Y. Chang, Y. Wu, Q. Wu, and L. Lin, ``MemCollab: Cross-Agent
Memory Collaboration via Contrastive Trajectory Distillation,''
arXiv:2603.23234, 2026.

[18] Z. Yang, W. Huan, L. Yingshuo, T. Haomiao, and L. Hujite, ``A
Heterogeneous Temporal Memory Governance Framework for Long-Term LLM
Persona Consistency,'' arXiv:2605.14802, 2026.

[19] K. Kim, M. Kang, T. Kim, Y. Yang, M. Ren, and S. J. Hwang,
``Memory Transfer Learning: How Memories are Transferred Across Domains
in Coding Agents,'' arXiv:2604.14004, 2026.

[20] K. Jiang, X. Cai, Z. Cui, A. Li, Y. Ren, H. Yu, H. Yang, D. Fu,
L. Wen, and P. Cai, ``KoMA: Knowledge-driven Multi-agent Framework for
Autonomous Driving with Large Language Models,'' arXiv:2407.14239, 2024;
IEEE Transactions on Intelligent Vehicles, 2024.

[21] A. Rezazadeh, Z. Li, A. Lou, Y. Zhao, W. Wei, and Y. Bao,
``Collaborative Memory: Multi-User Memory Sharing in LLM Agents with
Dynamic Access Control,'' arXiv:2505.18279, 2025.

[22] ``Contextual Orchestration and Scoped Memory Protocol with
Decentralized Memory Wallet for Adaptive Artificial Intelligence
Systems,'' U.S. patent filing, USPTO, accessed 2026.

[23] T. Helmi, ``Modeling Response Consistency in Multi-Agent LLM
Systems: A Comparative Analysis of Shared and Separate Context
Approaches,'' arXiv:2504.07303, 2025.

[24] S. Oram, O. Bhushan, and P. Bhattacharyya, ``Towards Emotion
Consistency Analysis of Large Language Models in Emotional
Conversational Contexts,'' arXiv:2605.06476, 2026.

[25] B. Asseri, E. Abdelaziz, M. Al Mogren, T. Alhefdhi, and A.
Al-Wabil, ``Deciphering Emotions in Children Storybooks: A Comparative
Analysis of Multimodal LLMs in Educational Applications,''
arXiv:2506.18201, 2025.

[26] ``Emotion in Context: Human and AI Perspectives on the
Affective Landscape of Severance,'' Corpus Pragmatics, Springer, 2025.

[27] C. Lam, J. Li, L. Zhang, and K. Zhao, ``Governing Evolving
Memory in LLM Agents: Risks, Mechanisms, and the Stability and Safety
Governed Memory (SSGM) Framework,'' arXiv:2603.11768, 2026.

[28] ``Continuum Memory Architectures for Long-Horizon LLM Agents,''
arXiv:2601.09913, 2026.

[29] ``ByteRover: Agent-Native Memory Through LLM-Curated
Hierarchical Context,'' arXiv:2604.01599, 2026.

[30] ``Graph-based Agent Memory: Taxonomy, Techniques, and
Applications,'' arXiv:2602.05665, 2026.

[31] ``Memory in LLM-based Multi-agent Systems: Mechanisms,
Challenges, and Collective Intelligence,'' TechRxiv preprint, 2026.

[32] H. Gao and Y. Zhang, ``INMS: Memory Sharing for Large Language
Model based Agents,'' arXiv:2404.09982, 2024.

[33] D. Zhang, W. Li, K. Song, J. Lu, G. Li, L. Yang, and S. Li,
``Memory in Large Language Models: Mechanisms, Evaluation and
Evolution,'' arXiv:2509.18868, 2025.

[34] C. Adam, A. Herzig, and D. Longin, “A logical formalization of the OCC theory of emotions,” Synthese, vol. 168, no. 2, pp. 201–248, 2009, doi: 10.1007/s11229-009-9460-9.

\end{document}